\documentclass{article}
\usepackage{IEEEtrantools}
\usepackage{amsmath,graphicx}
\usepackage{algorithm}
\usepackage{algorithmic}
\usepackage{array}
\usepackage{amsthm}
\usepackage{hyperref}

\title{CLASSP: a Biologically-Inspired Approach to Continual Learning through Adjustment Suppression and Sparsity Promotion}

\author{Oswaldo Ludwig}
\date{} 

\begin{document}

\maketitle

\begin{abstract}
This paper introduces a new biologically-inspired training method named \textbf{C}ontinual \textbf{L}earning through \textbf{A}djustment \textbf{S}uppression and \textbf{S}parsity \textbf{P}romotion (CLASSP). CLASSP is based on two main principles observed in neuroscience, particularly in the context of synaptic transmission and Long-Term Potentiation (LTP). The first principle is a decay rate over the weight adjustment, which is implemented as a generalization of the AdaGrad optimization algorithm. This means that weights that have received many updates should have lower learning rates as they likely encode important information about previously seen data. However, this principle results in a diffuse distribution of updates throughout the model, as it promotes updates for weights that haven’t been previously updated, while a sparse update distribution is preferred to leave weights unassigned for future tasks. Therefore, the second principle introduces a threshold on the loss gradient. This promotes sparse learning by updating a weight only if the loss gradient with respect to that weight is above a certain threshold, i.e. only updating weights with a significant impact on the current loss. Both principles reflect phenomena observed in LTP, where a threshold effect and a gradual saturation of potentiation have been observed. CLASSP is implemented in a Python/PyTorch class, making it applicable to any model. When compared with Elastic Weight Consolidation (EWC) using computer vision and sentiment analysis datasets, CLASSP demonstrates superior performance in terms of accuracy and memory footprint.
\end{abstract}

\section{Introduction}
\label{intro}

In the rapidly evolving field of machine learning, continuous learning has emerged as a critical area of investigation. Learning from a stream data, i.e. adapting to new information while retaining previously learned knowledge, is a fundamental aspect of intelligence. However, achieving this in artificial systems presents significant challenges due to the so-called catastrophic forgetting problem.

This paper introduces a new training method, called \textbf{C}ontinual \textbf{L}earning through \textbf{A}djustment \textbf{S}uppression and \textbf{S}parsity \textbf{P}romotion (CLASSP), inspired by principles observed in neuroscience. CLASSP addresses the challenges of continual learning by implementing a decay rate over the weight adjustment and a threshold on the loss gradient related to this weight. These principles aim to balance the need to learn new information and preserve important previously learned knowledge.

The paper is organized as follows. Section \ref{SOTA} briefly reports the state-of-the-art in Continual Learning and Catastrophic Forgetting, Section \ref{CLASSP} justifies and explains the proposed training algorithm. Section \ref{Experiments} shows the effectiveness of CLASSP through an ablation study and its comparison with AdaGrad, Adam, SGD and Elastic Weight Consolidation (EWC) methods using computer vision and sentiment analysis datasets. Conclusions are presented in Section \ref{Conclusions}.

\section{State-of-the-art Continual Learning}
\label{SOTA}

Continual learning, also known as lifelong or incremental learning, is a critical capability for intelligent systems to develop adaptively, acquiring, updating, accumulating and exploring knowledge incrementally over the time. The general objectives of continual learning are to ensure an appropriate balance between stability and plasticity and adequate intra/inter-task generalization \cite{ref1}.

The most common approaches to continuous learning include dynamic topology approaches, regularization approaches, and rehearsal (or pseudo-rehearsal) approaches \cite{ref3}. One of the most cited papers on regularized methods for continuous learning is the EWC method, which provides theoretical support for overcoming catastrophic forgetting in neural networks \cite{ref6} and is the base method in this study. EWC has been applied in practical scenarios for continuous learning of neural networks on diverse training sets, stabilizing the learning process and mitigating catastrophic forgetting.

Catastrophic forgetting occurs when learning a new task results in a dramatic degradation of performance on previously trained tasks, it's a significant problem for continual learning \cite{ref1}. The simplest and most common methods replay previous data during training, which violates the constraints of the ideal continuous learning configuration \cite{ref5}. However, recent advances have broadened our understanding and application of continuous learning with methods such as Relevance Mapping Network (RMN) \cite{ref5}, which is inspired by the Optimal Overlap Hypothesis, i.e. it aims for unrelated tasks to use distinct network parameters while allowing similar tasks to share some representation in order to significantly outperform data replay methods without violating the constraints for an ideal continuous learning system.

Both CLASSP and RMN consider the relevance of weights to previously learned data, but apply different strategies to achieve this. While CLASSP promotes a sparse distribution of updates throughout the model, leaving some weights unassigned for future tasks, RMN does not explicitly promote sparsity, but learns an optimal overlap of network parameters, which can result in a form of sparsity.

\section{CLASSP}
\label{CLASSP}

CLASSP leverages two key principles to address catastrophic forgetting. The first principle introduces a decay rate during weight updates. This implies that weights which have undergone numerous updates should receive lower learning rates. This is due to the likelihood that these weights encode crucial information relating to previously learned data. However, this principle leads to a diffuse distribution of updates throughout the model, as it encourages updates to weights that have not been previously updated. A sparse distribution of updates is more desirable in order to reserve weights for future tasks. To address this, the second principle introduces a threshold on the loss gradient. This promotes sparsity by only updating weights with a significant impact on the current loss, effectively reserving capacity for future tasks.

Both principles are observed in neuroscience, particularly in the context of synaptic transmission and Long-Term Potentiation (LTP). The concept of a decay rate over weight adjustment is akin to synaptic plasticity, where the strength of synaptic connections changes over time with experience. The study \cite{kopanitsa2006recording} explores the induction of LTP in hippocampal slices, demonstrating the threshold effect and saturation/decay of potentiation. Sparse learning is a concept where only a subset of weights are updated during training, which is similar to how certain synaptic connections are strengthened in the brain while others remain unchanged. The paper \cite{perez2021sparse} presents a sparse backpropagation algorithm for Spiking Neural Network (SNN), which can be seen as analogous to threshold-based updating in biological neurons.

Since the weight updates in backpropagation are directly proportional to the derivative of the loss function with respect to the weight, we leverage the previous derivative values directly within our decay term, rather than the previous weight updates, as shown in (\ref{CLASSP_update}). This makes our method a generalization of the AdaGrad algorithm \cite{duchi2011adaptive}, since AdaGrad is a special instance of CLASSP for $p=2$ and $threshold=0$. AdaGrad wasn’t specifically designed with continual learning in mind; was proposed as a method for adapting the learning rate for each parameter individually. However, it's interesting to note that some previous studies have found that AdaGrad can perform surprisingly well in certain continual learning scenarios \cite{hsu2019reevaluating}.

\begin{equation}
\label{CLASSP_update}
w_{i,t+1} = 
\begin{cases} 
w_{i,t} - \frac{\alpha \nabla L(w_{i,t})}{\sqrt[p]{\epsilon + \sum_{k=0}^{t-1}\left|\nabla L(w_{i,k})\right|^p}} & \text{if } \nabla L(w_{i,t})^2 > \text{threshold} \\
w_{i,t} & \text{otherwise}
\end{cases}
\end{equation}
where $\nabla L(w_{i,t})$ is the derivative of loss function $L$ with respect to the model parameter $w_i$ at iteration $t$ and $\epsilon$ is a small value used to avoid numerical issues.

Viewed through the lens of EWC, the first principle can be interpreted as applying a smaller learning rate to weights that cause a larger change in the loss function for previous tasks, i.e. larger squared derivatives of the loss with respect to the weight (the diagonal of the Fisher information matrix), indicating that these weights are likely more crucial to the performance on these tasks.

CLASSP\footnote{Code available here: \url{https://github.com/oswaldoludwig/CLASSP}} is detailed in Algorithm \ref{algo_CLASSP}, where we see that the CLASSP optimizer offers a more cost-effective approach to continual learning compared to EWC, since EWC requires buffering all the weights of the previous model and elements of the Fisher Information Matrix (FIM), which can be memory-intensive. On the other hand, CLASSP only needs to buffer one scaling factor per weight, which can significantly reduce the memory requirements and computational overhead. This makes CLASSP a potentially more scalable and efficient solution for continual learning scenarios.

\begin{algorithm}
  \caption{CLASSP Optimizer}
	\label{algo_CLASSP}
  \begin{algorithmic}[1]
     \REQUIRE{$params:$ learning rate $\alpha$, $threshold$, power $p$, $apply\_decay$ and $\epsilon$}
     \ENSURE{$loss$}
     \STATE{Initialize CLASSP with $\alpha$, $threshold$, power $p$, $apply\_decay$ and $\epsilon$}
     \FOR{each step in optimization}
     \STATE{Calculate $loss$ with autograd}
		 \STATE{Calculate $grad\leftarrow\nabla loss(w)$ with autograd for all parameters $w$}
     \FOR{each group of parameters}
     \FOR{each parameter $w$ in group}
     \IF{gradient of $w$ is not None}
     \STATE{Initialize $grad\_sum$ for $w$ if not already done}
		 \IF{$grad^2 > threshold$}
     \STATE{Update $grad\_sum$ for $w$:}
		 \STATE{$grad\_sum \leftarrow grad\_sum + \left|grad\right|^p$}
		 \IF{$apply\_decay$ is True}
		 \STATE{Calculate scaling factor for $w$:} 
		 \STATE{$scaling\_factor \leftarrow \alpha / \sqrt[p]{\epsilon + grad\_sum}$}
     \STATE{Update $w$: $w \leftarrow w - scaling\_factor * grad$}
     \ENDIF
		 \ENDIF
     \ENDIF
     \ENDFOR
     \ENDFOR
		 \ENDFOR
     \RETURN{$loss$}
  \end{algorithmic}
\end{algorithm}

\section{Experiments}
\label{Experiments}

This section presents the results of our experiments to evaluate the performance of CLASSP in the continual learning setting. The first set of experiments uses two popular computer vision datasets: MNIST \cite{lecun1998gradient} and Fashion MNIST (FMNIST) \cite{xiao2017fashion}. The second set of experiments uses two sentiment analysis datasets: Financial Phrasebank (FPB) \cite{malo2014good} and IMDB \cite{maas2011learning}. To isolate the issue of forgetting from the model's capacity to generalize, the same datasets were used for both training and evaluation.

\subsection{Computer Vision}
\label{CV}

\begin{figure}[H]
\vskip -0.0in
\begin{center}
\centerline{\includegraphics[width=1.0\columnwidth]{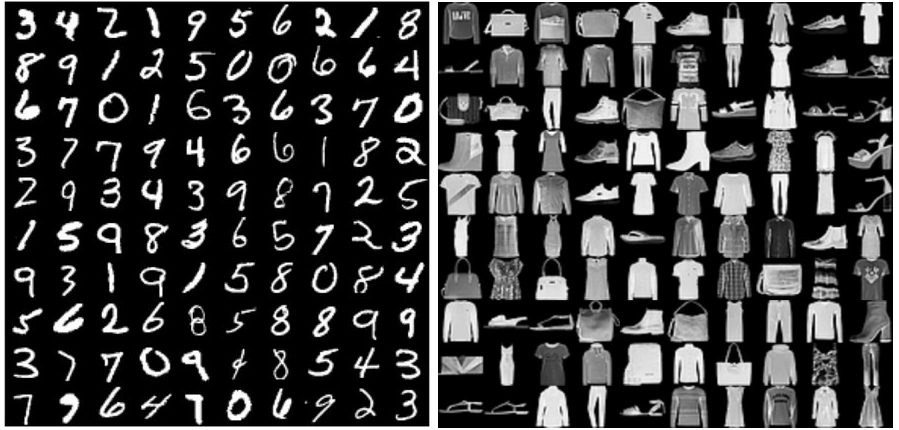}}
\caption{Samples from MNIST (left) and FMNIST (right).}
\label{datasets}
\end{center}
\vskip 0.0in
\end{figure}

For all computer vision experiments, the model is trained for 4 epochs on MNIST, followed by 1 epoch on FMNIST. This setup allows assessment of how well the model retains knowledge of MNIST after learning FMNIST. Figure \ref{datasets} shows the difference between the two datasets.

The model architecture leverages a Convolutional Neural Network (CNN) implemented in PyTorch\footnote{See the script for the experiment with CLASSP in computer vision at \url{https://github.com/oswaldoludwig/CLASSP/blob/main/experiment_CV.py}}. The CNN comprises two convolutional layers, each with 3x3 filters, interspersed with dropout layers for regularization, and two fully-connected layers for final classification. The first convolutional layer applies 32 filters, followed by a ReLU activation function. The second convolutional layer utilizes 64 filters and is followed by a ReLU activation and a max pooling operation. The loss function is cross-entropy for all experiments except for EWC\footnote{See the EWC script in computer vision at \url{https://github.com/oswaldoludwig/CLASSP/blob/main/experiment_EWC.py}}. 

The results summarized in Table 1 provide a comparative analysis of various training algorithms on the sequence of MNIST and FMNIST datasets. A series of 10 experiments per algorithm were conducted, from which the mean accuracy and standard deviation were derived.

The best configuration for CLASSP was established with learning rate of 0.2, power $p=1$ and $threshold$ of 0.5, which was applied only to the first dataset, to leave unassigned weights for the second tasks. The decay rate was applied exclusively to the second dataset, i.e. the optimizer computes the sum of the gradients while processing the first dataset, but applies the decay rate only to the second dataset. This process is detailed in Lines 10-15 of Algorithm \ref{algo_CLASSP}. 

Table 1 also presents an ablation study. The second row shows the performance of CLASSP without threshold, while the third row shows how CLASSP performs as a conventional AdaGrad optimizer, i.e. with power $p=2$, without threshold, and the decay rate applied to both datasets.

The last two rows of Table 1 display the accuracies for both the vanilla Stochastic Gradient Descent (SGD) and EWC, which serve as baseline algorithms. 

\begin{table}[h]
\label{table_results}
\centering
\caption{Comparative analysis of different training algorithms on the dataset sequence MNIST + FMNIST: mean accuracy and standard deviation calculated over 10 experiments.}
\begin{tabular}{|l|l|r|r|}
\hline
\textbf{training algorithm} & \textbf{dataset} & \textbf{mean} & \textbf{std} \\ \hline
CLASSP & MNIST & 91.05 & 1.94 \\
\scriptsize{$\alpha=0.2$, threshold=0.5, decay rate on first dataset=False, p=1} & FMNIST & 85.53 & 0.83 \\ \hline
CLASSP without threshold & MNIST & 80.56 & 1.55 \\
\scriptsize{$\alpha=0.2$, threshold=0.0, decay rate on first dataset=False, p=1} & FMNIST & 85.76 & 2.19 \\ \hline
CLASSP (AdaGrad instance) & MNIST & 66.10 & 3.09 \\
\scriptsize{$\alpha=0.2$, threshold=0.0, decay rate on first dataset=True, \textbf{p=2}} & FMNIST & 81.69 & 0.91 \\ \hline
Vanilla SGD & MNIST & 43.18 & 0.90 \\
   & FMNIST & 85.16 & 0.64 \\ \hline
EWC & MNIST & 68.52 & 5.32 \\
   & FMNIST & 82.48 & 1.19 \\ \hline
\end{tabular}
\end{table}

The ablation study further highlights the impact of CLASSP's unique features. When CLASSP is reduced to a conventional AdaGrad optimizer, the mean accuracy drops to $66.10\%$ on MNIST, underscoring the benefits of the custom decay rate, adjustable power $p$, and threshold mechanism.

In comparison, Vanilla SGD and EWC show lower accuracies and, in the case of EWC, a higher standard deviation, suggesting less stable performance across experiments. This reinforces the value of CLASSP approach, which not only enhances learning outcomes but also provides more consistent results.

\subsection{Text Classification}
\label{SA}

In all sentiment analysis experiments, the FPB dataset is used for initial training, followed by further training on the IMDB dataset\footnote{See the script for the experiment with CLASSP in sentiment analysis at \url{https://github.com/oswaldoludwig/CLASSP/blob/main/experiment_SA.py}}. Both datasets involve sentiment analysis, but differ in context: FPB focuses on financial news articles, aiming to predict the impact on stock prices, while IMDB consists of movie reviews, providing sentiment analysis for the entertainment domain.

The experiments leverage a Transformer Encoder for text classification. The model first embeds the input text using a 600-dimensional embedding layer. Dropout with a rate of 0.1 and layer normalization are then applied. The processed signal is then fed into a single-layer Transformer encoder with 30 attention heads. Finally, the output is normalized and fed into a linear layer with three units, corresponding to the three sentiment classes (positive, neutral, negative) based on the FPB annotation scheme.

To prevent overfitting on the IMDB dataset, which could worsen forgetting of the FPB dataset, we employ a threshold on the loss function during IMDB training. This threshold, which is a function of the optimizer, maintains IMDB accuracy at approximately $85\%$ ensuring a fair comparison. While all optimizers achieved over $99\%$ accuracy on FPB, CLASSP exhibited superior performance in mitigating forgetting. The forgetting rate observed with CLASSP ($33.34\%$) was significantly lower compared to Adam ($53.09\%$) or even EWC\footnote{See the EWC script in sentiment analysis at \url{https://github.com/oswaldoludwig/CLASSP/blob/main/experiment_SA_EWC.py}} ($47.15\%$).

To understand the contribution of different components in CLASSP, Table 2 presents the results of an ablation study, alongside baseline performance of SGD, Adam, and EWC.

\begin{table}[h]
\label{table_results_2}
\centering
\caption{Comparative analysis of different training algorithms on the dataset sequence FPB + IMDB: mean accuracy and standard deviation calculated over 10 experiments.}
\begin{tabular}{|l|l|r|r|}
\hline
\textbf{training algorithm} & \textbf{dataset} & \textbf{mean} & \textbf{std} \\ \hline
CLASSP & FPB & 66.04 & 3.85 \\
\scriptsize{$\alpha=0.025$, threshold=0.99, p=1} & IMDB & 85.74 & 1.30 \\ \hline
CLASSP without threshold & FPB & 64.81 & 6.63 \\
\scriptsize{$\alpha=0.025$, threshold=0.0, p=1} & IMDB & 85.48 & 1.38 \\ \hline
CLASSP (AdaGrad instance) & FPB & 52.95 & 1.97 \\
\scriptsize{$\alpha=0.025$, threshold=0.0, \textbf{p=2}} & IMDB & 85.01 & 2.53 \\ \hline
Vanilla SGD & FPB & 47.24 & 5.57 \\
   & IMDB & 85.67 & 1.72 \\ \hline
Adam & FPB & 46.63 & 3.72 \\
   & IMDB & 85.66 & 1.99 \\ \hline
EWC & FPB & 52.46 & 1.49 \\
   & IMDB & 85.67 & 1.15 \\ \hline
\end{tabular}
\end{table}

Once again $p=1$ is the optimal power. The power $p$ in both Adagrad and CLASSP algorithms determines how the past gradients are accumulated and used to adjust the learning rate. In Adagrad, $p=2$ means that larger gradients have a stronger influence in decreasing the learning rate, as they are squared before being accumulated. This can lead to a rapid decrease in the learning rate, especially if large gradients are encountered early in training, as usual. On the other hand, CLASSP uses the sum of absolute values of past gradients, i.e. $p=1$. This means that all gradients, regardless of their magnitude, contribute linearly to the accumulated sum, leading to a more balanced and slower decrease of the learning rate, as it is less sensitive to large gradients.

In summary, these experiments encompassing both computer vision and sentiment analysis tasks demonstrate the effectiveness of CLASSP in mitigating catastrophic forgetting during continual learning.

\section{Conclusions}
\label{Conclusions}

This paper introduces CLASSP, a novel continual learning method inspired by biological learning principles. CLASSP tackles catastrophic forgetting by balancing the acquisition of new information with the preservation of past knowledge. This is achieved through two key mechanisms: a decay rate on weight updates and a threshold on the loss gradient. Decay rate assigns smaller learning rates to weights that have been frequently updated, thus preserving their relevance to past tasks, while the threshold promotes sparsity, reserving capacity for future tasks.

Compared to existing methods like EWC, CLASSP offers advantages in terms of memory footprint and performance. It only requires storing weight-specific scaling factors, as opposed to EWC's need to buffer all past weights and elements of the Fisher Information Matrix. Experiments on MNIST, Fashion-MNIST, IMDB and Financial Phrasebank datasets demonstrate CLASSP's effectiveness, achieving higher average accuracy compared to AdaGrad, SGD, Adam and EWC, validating CLASSP as a promising method for continual learning, with potential applications in various domains where learning from sequential or streaming data is crucial.

As future work, I see promise in applying CLASSP to more complex datasets and tasks beyond computer vision and text classification. Investigating the impact of scale factor quantization for further memory efficiency is also of interest. Finally, it would be valuable to delve deeper into the theoretical aspects of CLASSP's convergence properties. This could, for example, start by examining Theorem 6 of \cite{wang2023convergence}, taking into account the added complexity introduced by the thresholding mechanism, to gain a better understanding of the interplay between the adaptive learning rate and the thresholding mechanism.

\bibliographystyle{IEEEtran}

\bibliography{biblio}

\end{document}